# Measuring Competency of Machine Learning Systems and Enforcing Reliability

Michael Planer[1], Jen Sierchio[2],

BAE Systems, Inc. michael.planer@baesystems.com[1], jennifer.sierchio@baesystems.com[2]

**Abstract**

We explore the impact of environmental conditions on the competency of machine learning agents and how real-time competency assessments improve the reliability of ML agents. We learn a representation of conditions which impact the strategies and performance of the ML agent enabling determination of actions the agent can make to maintain operator expectations in the case of a convolutional neural network that leverages visual imagery to aid in the obstacle avoidance task of a simulated self-driving vehicle.

## Introduction

Opaque Machine Learning (ML) models (Random Forest, Neural Networks, etc) lack intrinsic interpretability (Belle 2021), their output comes without a clear estimate of their expected accuracy (Szegedy 2014), and they often encounter situations beyond the scope of their training (Xu 2020). We assess the competency of a convolutional neural network (CNN) trained to identify relative positions of a variety of obstacles to aid in navigation of a self-driving car.

Monitoring competency can help to address the above issues by learning a representation of the input data that is compact and can be compared quickly to identify when incoming data is sufficiently different from training data samples. This representation (conditions) can be used as input to train a classifier or regression that predicts competency of the system. By comparing these conditions across competency regimes, it is possible to succinctly explain what drives the competency of the ML agent. In our CNN example, the CAML agent may specify that conditions associated with "darkness" may cause the agent to miscalculate the distance to the embankment ahead.

## Competency of a Machine Learning Agent

When a machine learning agent completes a task, there are several aspects of its competency that can be considered. Often one is concerned with a performance metric (typically accuracy) of an ML model. For the case of estimating distance to obstacles, various metrics may be relevant. For the purposes of discussion, we can use the Mean Squared Error (MSE), but the conclusions presented can extend to more complex performance metrics (Sierchio 2022).

Another aspect of competency is "strategy"; this is a generalization of the precise behavior the ML agent used to complete its task. For this use case, the trace of the activation patterns given an input image is treated as a behavior, and a grouping of similar activation patterns is considered the strategy (Planer 2021). If a specific set of input conditions is normally tied to a particular strategy (e.g., "darkness" is typically tied to strategy 1 in our training data), but during online operation another strategy is used (e.g., strategy 8 is only associated with "sunny"), the ML agent is likely using an inappropriate strategy for the current input image. This can be extrapolated to other ML tasks such as deep reinforcement learning (Filiberti 2021).

## Learning and Predicting Competency of a Machine Learning agent

In order to estimate the competency of an ML agent, an ML agent based on the Alexnet CNN (Krizhevsky 2012) was trained to identify the distance to obstacles to aid in collision avoidance; obstacles within 8 meters required a maneuver. 80,000 images were generated for a variety of environmental conditions (including 'rain', 'snow', 'dusk', 'night') using the Gazebo simulation environment. The agent was trained on half of the images. Figure 1 shows a few examples from training dataset.

The competency of the ML agent was measured on the training dataset, leveraging hierarchical Dirichlet processes (HDPs) to generate a topic distribution that represents competency controlling conditions. Inclusion of competency information in the form of strategy and performance measurements during the learning process of the HDPs enables generation of a much more useful set of conditions. Details of the CAML system and evaluation application can be found in presentations by Dean Schifilliti (2022) and Michael Planer (2021).

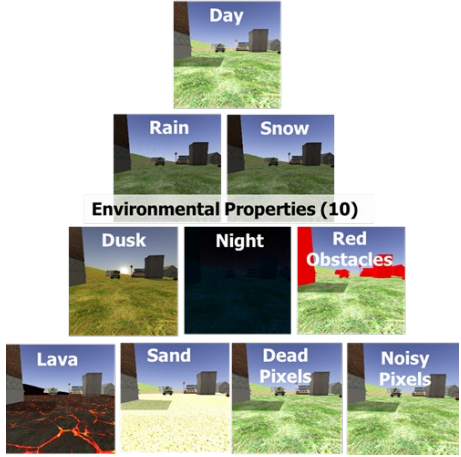

Figure 1. Set of environmental parameters and example images as input to the CNN during the training of the ML agent (Schifilliti 2021).

Performance and strategy were predicted by models that take those learned conditions as input. These estimates inform a user of the likely competency of the ML agent under the given conditions, and can provide automatic suggestions to maintain more reliable operation.

## Assessing the Capability of the Competency Aware System

Evaluation of the described competency system was presented by Jen Sierchio (2022) and Dean Schifilliti (2022).
<u>Coverage</u> measures the ability of the CAML system to correctly identify competency controlling conditions. In the case of the above system in Year 2 of the DARPA CAML program, we achieved >95% coverage.
<u>Correctness</u> reports the ratio of times the CAML predicted strategy was correctly predicted to the total number of trials. This approach achieved 90% correctness when estimating the single strategy that would be executed. When estimating the likely strategy distribution given the competency controlling conditions (taking into account the ML agent's intrinsic variability), this correctness metric rose to 100%.
<u>Fidelity</u> verifies how often a given estimate falls within the expected performance band given a set of conditions with the Brier Score. Our approach scored an 80% Fidelity when constrained to a high-level summary over only a few combinations of conditions. But when allowed the freedom to assess each measurement using the full complexity of the HDP derived conditions, the fidelity score rose to 99%.
<u>Reliability</u> is a measure of how often the ML agent meets or exceeds a set of operator defined requirements. In the case of the CNN distance estimator, an example could be: "when sunny, 99% of obstacles must be detected in time to avoid them". The CAML system has a representation of conditions and can identify when the "sunny" condition is present. It then leverages its estimate of performance to estimate the likelihood the CNN will predict an obstacle that is close as far enough away to continue on the vehicle's current trajectory. In the case where the CNN is likely to fail is greater than 99%, the CAML agent suggests an alternative to the operator or ML agent itself.

Our research into competency aware machine learning continues with a focus on improving reliability and extension of this approach to facilitate control over swarms of UASs.

## Acknowledgments

This work was funded through the DARPA Competency Aware Machine Learning (CAML) program.